\documentclass[runningheads]{llncs}
\usepackage[T1]{fontenc}
\usepackage{booktabs}
\usepackage{graphicx}
\usepackage{hyperref}
\usepackage{subfig}

\begin{document}
\title{A Few-shot Approach to Resume Information Extraction via Prompts}
%
%

\author{Chengguang Gan\orcidID{0000-0001-8034-0993} \and
Tatsunori Mori\orcidID{0000-0003-0656-6518}}

\authorrunning{Gan, Mori}

\institute{Yokohama National University, Japan\\
\email{gan-chengguan-pw@ynu.jp}, \email{tmori@ynu.ac.jp}}

\maketitle              

\begin{abstract}
Prompt learning's fine-tune performance on text classification tasks has attracted the NLP community.  This paper applies it to resume information extraction, improving existing methods for this task.  We created manual templates and verbalizers tailored to resume texts and compared the performance of Masked Language Model (MLM) and Seq2Seq PLMs.  Also, we enhanced the verbalizer design for Knowledgeable Prompt-tuning, contributing to prompt template design across NLP tasks.
We present the Manual Knowledgeable Verbalizer (MKV), a rule for constructing verbalizers for specific applications.  Our tests show that MKV rules yield more effective, robust templates and verbalizers than existing methods.  Our MKV approach resolved sample imbalance, surpassing current automatic prompt methods.  This study underscores the value of tailored prompt learning for resume extraction, stressing the importance of custom-designed templates and verbalizers.

\end{abstract}
\section{Introduction}

With the introduction of the Transformer architecture \cite{vaswani2017attention}, large-scale language models pre-trained on unsupervised tasks have consistently achieved state-of-the-art results on a wide range of NLP tasks. The most prominent of these models include the Encoder-based BERT \cite{devlin-etal-2019-bert}, a task-based MLM pre-trained model primarily used for classification tasks, and its modified version, RoBERTa \cite{liu2019roberta}. Another example is the T5 model \cite{raffel2020exploring}, which uses the Seq2Seq MLM pre-training method. The pre-training and fine-tuning paradigm was widely applied to various NLP downstream tasks until 2020, when a novel pre-training and prompt paradigm was proposed.
\par The first prompt-based approach for text classification tasks, which utilizes models to answer cloze questions and predict labels, was proposed for sentiment classification \cite{schick-schutze-2021-exploiting}. This approach was shown to achieve higher accuracy than the traditional fine-tuning paradigm, even with limited training data. The underlying principle of prompt learning involves using manually designed templates to wrap sentences, which are then masked to provide the target label relation words. This masked text is then input into the model, which predicts the corresponding label relation words. The development of manual prompt templates led to the exploration of automatic prompt generation methods. These methods are broadly categorized into two groups: discrete prompts and continuous prompts \cite{liu2023pre}.
\par As previously noted, prompt methods have demonstrated exceptional performance across various benchmark datasets. However, their efficacy in unique practical scenarios, such as information extraction from resumes, remains a question. This application is of great significance to businesses, given the daily influx of resumes. Though deep learning has facilitated automated resume screening, the resource demand for annotating resume texts for Pretrained Language Model (PLM) fine-tuning can be prohibitive. Prompt methods, requiring only a few labeled resume texts, present an economical solution, particularly for small businesses or niche industries. In this study, we leverage a seven-category sentence classification of English resume data~\cite{gan2022construction}. This approach transforms the resume extraction task into a sentence classification task, aligns with the current research focus in prompt learning, and offers practical value, especially for organizations with limited resources. Moreover, the recurring specific words in resumes are ideal for constructing a Knowledgeable Verbalizer (KV), a technique known for its state-of-the-art (SOTA) results~\cite{hu-etal-2022-knowledgeable}. Thus, we focus on information extraction from resumes in this study.

\begin{figure*}[!t]
\centering
\includegraphics[width=347.25 pt]{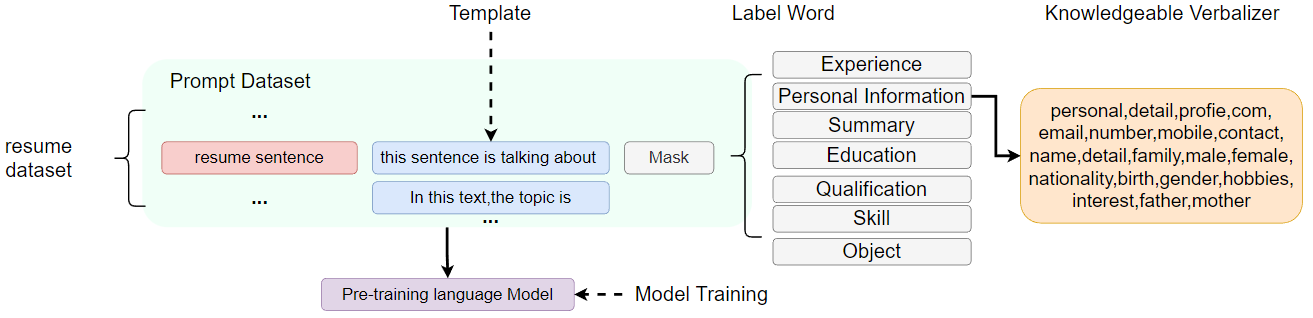}
\caption{\label{figure1}Illustration of the process of prompt learning for the resume information extraction task.}

\end{figure*}

\par As depicted in Figure \ref{figure1}, the sentence to be classified, denoted as X, is inputted into the wrap class. Following this, an indicative template sentence is appended. Subsequently, a mask token is integrated into the prompted sentence, representing the label relation word to be forecasted, such as “$\{$X$\}$ this sentence is about $\{$mask$\}$”. Consequently, a sentence classification problem is reframed as a fill-in-the-blanks (mask token) task. The Masked Language Model (MLM) explicitly addresses this issue. As a result, prompting enables the downstream task to align more closely with the pre-training task, facilitating rapid model adaptation for small datasets. The Knowledgeable Verbalizer (KV) extends a single label relation word to cover multiple associated terms. Initially, the output probability of each relevant word for the mask token is calculated. These probabilities are then consolidated, and the class with the highest probability of associated words is selected as the sentence's predicted outcome, i.e., the mask token word. In conclusion, KV with a manually designed template has achieved superior scores across numerous datasets compared to other prompt methods' baseline scores. Therefore, our initial hypothesis in this study was that the amalgamation of manually designed templates and KV is the most effective among current mainstream prompt methods. We conducted subsequent comparison experiments to validate this hypothesis. To assess the impact of different templates on the results, we devised several alternative templates for comparison experiments. The results suggest that the efficacy of prompt-learning is significantly impacted by template design. A baseline KV was created for resume text classification using the original Knowledgeable Prompt-learning method \cite{hu-etal-2022-knowledgeable} and compared with the KV designed based on our refined rules. We term the KV constructed following our proposed rule as Manual Knowledgeable Verbalizer (MKV). On the 25-shot and 50-shot tasks, performance increased significantly from 54.96$\%$ to 63.65$\%$ and 59.72$\%$ to 76.53$\%$ respectively. These findings indicate our proposed MKV is more suited for the resume extraction task. We conducted comparative experiments on two models using 25/50/10-shot tasks and two distinct training techniques to compare the encoder structure of BERT and the Seq2Seq structure of the T5 model in the context of prompt-learning. In summary, our primary contributions include:
\par 1. Development of a comprehensive set of prompt-learning techniques for the few-shot resume information extraction task.
\par 2. Proposal of KV construction rules, based on the original Knowledgeable Prompt-tuning (KPT), that are more compatible with resume text. This provides insights for subsequent researchers to develop KVs for practical application scenarios.
\par 3. Comparison of manual template construction with automatic template generation for the prompt method. We have demonstrated that, in the current state of prompt-learning for resume text, a method employing manually crafted prompts surpasses one using automatically generated prompts.

\section{Related Work}

The research related to this paper is divided into two main groups. Prompt templates and constructs of prompt verbalizer.
\par \textbf{Prompt Template}. Since PET(Pattern-Exploiting Training)\cite{schick-schutze-2021-exploiting} was proposed, a surge of prompt research has been started. A PET follow-up study mentioned that smaller size models, such as BERT-base, are also capable of few-shot learning\cite{schick-schutze-2021-just}. There are also some other studies of discrete prompts\cite{gao-etal-2021-making}\cite{reynolds2021prompt}\cite{tam-etal-2021-improving}. The idea behind a discrete prompt is that the words in the template are real. The reason for this is because in coordinate space, the words are in a discrete state. Another hand, as all token-level prompts have been automatically generated, some research has gone a step further by replacing the token-level prompt template in the token representation with continuous vectors directly and training these prompts instead of the fine-tune model(A.K.A Continuous Prompt)\cite{li-liang-2021-prefix}\cite{lester-etal-2021-power}\cite{qin-eisner-2021-learning}\cite{hambardzumyan-etal-2021-warp}\cite{liu-etal-2022-p}. Using continuous vectors, we can find the best set of vectors to replace the discrete prompt template.
\par \textbf{Prompt Verbalizer}. The automatic construction method of verbalizer in few-shot learning is explored\cite{schick-etal-2020-automatically}. There are also corresponding verbalizers for some of the prompt methods mentioned in the previous paragraph, such as manual, soft and automatic verbalizers. Finally, the KV in this study also has better performance than other verbalizers\cite{hu-etal-2022-knowledgeable}.

\section{Task Setup}

In this work, we select the resume information sentence classification dataset as the experimental object. Hence, this study also considers the task setting from a practical application scenario. Suppose an IT company needs to fine-tune its company resume information extraction model. First, a training dataset consisting of several hundred resumes needs to be constructed. This results in the need to annotate tens of thousands of sentences. This is a non-negligible cost for a small company. Also, some small start-up companies may not have hundreds of resumes to create a training dataset. Thus, it is essential to minimize the investment of company human resource in producing training datasets for company.
\par In summary, we set the task to be in the case of annotating only one or two resumes. The resume dataset was extracted from 15,000 original resumes and 1000 of them were used as tagged objects \cite{gan2022construction} \footnote{\raggedright\url{https://www.kaggle.com/datasets/oo7kartik/resume-text-batch}}. In addition, the total number of sentence samples in this dataset is 78786, and the total number of annotated resumes is 1000. So it is calculated that each resume contains about 79 sentences on average.

\section{Prompt Design}

In this section, we firstly designed a series of prompt templates for resumes. Secondly, we presented a different KV construction method from the original KPT depending on the textual characteristics of the resume(Figure \ref{figure1}).

\subsection{Manual Template}

For the manually designed templates, we divided into two design thoughts. One is the generic template that is commonly used in prior studies (e.g., “$\{$input sentence$\}$ In this sentence, the topic is $\{$mask$\}$.” ). Another type of template is designed for resume documents (e.g., “$\{$input sentence$\}$ this sentence belongs in the $\{$mask$\}$ section of the resume.”). Based on this, we designed a series of templates specifically for the resume text. By inserting words like “resume” and “curriculum vitae” into the templates, which are strongly related to the classification text. It is anticipated that the performance of the few-shot learning of resume text would improve with the specific design of templates and MKV.

\subsection{Knowledgeable Verbalizer}
\label{4.2}
In this study, KV from the KPT paper is utilized\cite{hu-etal-2022-knowledgeable}. In their study, KV is constructed by introducing external knowledge. As an example, consider the class label “experience”. By using related word search sites, the words that appear frequently in conjunction with “experience” are tallied, and the top 100 words with the highest frequency are selected to constitute KV\footnote{https://relatedwords.org/}. There are seven word sets for class labels to construct the KV.
\par
\begin{figure}[!h]
\centering
\includegraphics[width=219 pt]{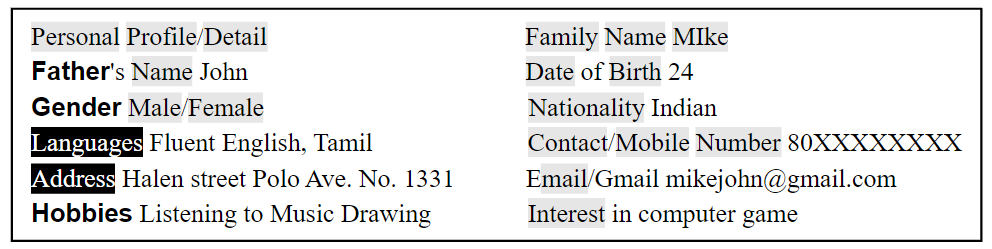}
\caption{\label{figure2}Example of label-related word selection for the \textbf{\emph{personal information}} class in resume when constructing Knowledgeable Verbalizer\protect\footnotemark.}

\end{figure}

 As shown in Figure \ref{figure2}, the text of the “personal information” in the original manuscript of a resume was selected to illustrate the MKV construction rules of our proposal. In summary, we present two rules for selecting the label relation word. \textbf{1}. Words that frequently appear in sentence of the target class with resume text. \textbf{2}. This word does not often occur in other classes.Thus, the words marked with gray background in the figure can be selected according to the two rules mentioned above. Two of them, “languages” and “address”, are marked with a black background. The word “languages” often appears in “skill” classes for programming languages.
The “address” is a word that appears not only in personal information as a home address. It also appears in “experience” classes for company addresses. So the above two words will not be selected in the label relation word set of personal information.

\footnotetext{Although the data comes from open source websites, mosaic is given to the part involving privacy.}

\section{Experiments Setup}

This study aims to investigate the impact of various factors on the outcomes of limited opportunity resume learning, as well as to assess the efficacy of our proposed MKV construction rules for Prompt Learning of resume material. We performed multiple comparative tests between the prompt's template and verbalizer, evaluated the performance of several prompting methods against MKV, and examined the effectiveness of two structurally distinct PLM models in few-shot resume learning. These experiments utilized a specially created resume dataset.
\par For efficient iteration of different experiments, we used Openprompt\footnote{https://github.com/thunlp/OpenPrompt}, an Open-Source Framework for Prompt-learning\cite{ding-etal-2022-openprompt}. We used the F1-micro as the evaluation metric for all experiments. Given that few-shot learning typically allows only a limited number of training samples per class, we employed a random seed to extract the training set from our unbalanced dataset of resumes. This approach maintained the original sample distribution, thereby preserving the inherent imbalance of the resume samples. Distinct random seeds were used for the 25/50/100-shot experiments.
\par Initially, we conducted two comparison experiments to determine the most efficient template and Knowledge Verbalizer (KV). One experiment compared different manual templates, while the other compared the performance of our proposed MKV construction method with the original KV construction method. After establishing the performance of Manual Template (MT) and Manual Knowledge Verbalizer (MKV), they were compared with three other types of templates and verbalizers.
\par The methods we considered include: 1) The Automatic method, where the model generates discrete prompt template words automatically, 2) Soft prompt method, where by optimizing the vectors of the embedding layer, soft prompts are utilized, 3) P-tuning method, where token-level templates are replaced with dense vectors, which are trained to predict the masked words, and 4) Prefix prompt approach, where task-specific continuous sequence prefixes are trained instead of the entire transformer model.
\par Finally, to investigate the fine-tuning and prompt-tuning performance of RoBERTa${\mbox{\scriptsize large}}$ and T5${\mbox{\scriptsize large}}$, we conducted an additional experiment. The scores from the first three experiments were obtained after 4 training epochs, while the PLM comparison experiment used scores adjusted to the best epoch based on test set scores. To evaluate the performance of KV and MKV in the context of sample imbalance, we created and analyzed the confusion matrix of the 50-shot test dataset for both methods.

\begin{table*}[!h]
\centering
\caption{\label{table1}
The result of 0-shot was compared with different Template. Use the manual template(MT) and Manual knowledgeable verbalizer(MKV).}
\begin{tabular}{ccc}
\toprule[2pt]
\textbf{Method} & \textbf{Template} & \textbf{F1-score}\\
\midrule
& $\{$input sentence$\}$ In this sentence, the topic is $\{$mask$\}$. & 33.45 \\
\textbf{MT+MKV} & $\{$input sentence$\}$ this sentence is talking about $\{$mask$\}$. & 55.77 \\
&$\{$input sentence$\}$this sentence belongs in the $\{$mask$\}$  \\
& section of the resume. & 61.32 \\
&$\{$input sentence$\}$this sentence belongs in the $\{$mask$\}$  &  \\
& section of the curriculum vitae. & \textbf{62.09} \\
\bottomrule[2pt]
\end{tabular}
\end{table*}

\section{Results and Analysis}

\subsection{Comparison Between Different Manual Templates}

Table \ref{table1} showcases experimental results from comparing four distinct templates. To eliminate confounding variables such as training samples, we adopted a 0-shot training strategy, directly predicting the test set using the original parameters of PLMs. This amplifies the impact of each template's efficacy. The top two are universal prompt templates suitable for any classification task. Contrarily, we developed two templates specifically for the resume dataset; incorporating “resume” into the templates notably enhanced the outcomes.
\par Another noteworthy observation is the 0.77 point score increment following the substitution of “resume” with “curriculum vitae”. We hypothesize this arises from the polysemous nature of “resume” (e.g., n.summary, v.recover), creating ambiguity when used within the template sentence. Replacing “resume” with the unambiguous term “curriculum vitae” consequently improved the score.

\begin{table*}[!ht]
\centering
\caption{\label{table2}
The results of 0/50/100-shot was compared with different Knowledgeable Verbalizer construction method.}
\begin{tabular}{cccc}
\toprule[2pt]
\textbf{Total Label Set Size(KV method)} & \textbf{25-shot} & \textbf{50-shot} & \textbf{100-shot}\\
\midrule
700(KV-baseline)& 54.96 & 59.72 & 66.46 \\
63(Ours MKV) & \textbf{63.65} & \textbf{76.53} & \textbf{76.72} \\
\bottomrule[2pt]
\end{tabular}
\end{table*}

\subsection{Compare Different Verbalizers}

Subsequently, we compare the performance of the original KV construction method and our proposed method. As shown in Table \ref{table2}, the total label set size means the sum of the extended words of the seven categories. To start with, for baseline, we follow the method in the original paper to obtain a set of related words to each sentence class by retrieving webpages with the class label as a query\cite{hu-etal-2022-knowledgeable}. Later, we constructed a total of 63 MKVs according to the rule proposed in \ref{4.2}. For the KV comparison test in this section, we used the fourth manual template in Table \ref{table1} ($\{$text$\}$ this sentence belongs in the $\{$mask$\}$ section of the curriculum vitae.). MKV is 8.69/16.81/10.26 point higher than the origin KV on 25/50/100-shot. This result also demonstrates the effectiveness of our improved MKV.

\subsection{Comparison of Different Prompt Methods}

\begin{table*}[!ht]
\centering
\caption{\label{table3}
Compare the results of different prompt methods at 25/50/100-shot. MT, MV is the abbreviation of Manual Template and Verbalizer. ST, SV is the abbreviation of Soft Template and Verbalizer. PT, PFT is the abbreviation of P-tuning Template and Prefix-tuning Template. AutoV is the abbreviation of AutomaticVerbalizer. MKV is the abbreviation of Manual Knowledgeable Verbalizer.}
\begin{tabular}{ccccc}
\toprule[2pt]
\textbf{Prompt Method} &   &  & \textbf{F1-score} & \\
\midrule
Template & Verbalizer & 25-shot & 50-shot & 100-shot \\
\midrule
 & MKV & 63.65 & \textbf{76.53} & \textbf{76.72} \\
MT & MV & 51.95 & 70.18 & 72.89 \\
 & SV & 51.49 & 55.37 & 64.29 \\
 & AutoV & 14.72 & 14.48 & 13.93 \\
\midrule
 & MKV & \textbf{63.92} & 70.42 & 73.91 \\
 ST& MV & 39.93 & 57.91 & 65.42 \\
 & SV & 33.69 & 53.60 & 60.29 \\
 & AutoV & 14.72 & 14.48 & 13.93 \\
\midrule
 & MKV & 62.82 & 68.19 & 72.33 \\
 PT& MV & 42.11 & 55.97 & 65.65 \\
 & SV & 16.43 & 50.42 & 64.04 \\
 & AutoV & 14.48 & 14.28 & 15.12 \\
\midrule
 & MKV & 60.88 & 67.29 & 73.60 \\
 PFT& MV & 40.36 & 55.62 & 62.73 \\
 & SV & 39.13 & 53.56 & 57.62 \\
 & AutoV & 14.85 & 14.65 & 13.96 \\
\bottomrule[2pt]
\end{tabular}
\end{table*}

Since the OpenPrompt framework divides the process of prompt-tuning into two main parts: Template and Verbalizer (See in Figure \ref{figure1}), these two parts can be used in combination at will. Hence, we have selected four representative Templates and four Verbalizer methods and compared their few-shot learning effectiveness with each other. As shown in Table \ref{table3}, In the 50/100-shot experiments, the combination of MT+MKV achieved the best results. Especially in the 50-shot experiment, the combination of MT+MKV scored 6.11 point higher than the second place ST+MKV. However, in the 25-shot experiment, the ST+MKV combination was slightly higher than the MT+MKV combination. Overall, the MT+MKV method that we propose outperforms the other four template and verbalizer combinations. This further validates the effectiveness and robustness of our MKV constructed for the resume classification dataset.

\subsection{Results on T5 and RoBERTa Model}

\begin{table*}[!ht]
\centering

\caption{\label{table4}
Compare the few-shot(25/50/100-shot) learning results of RoBERTa$_{\mbox{\scriptsize large}}$ and T5$_{\mbox{\scriptsize large}}$ models under different methods.}
\begin{tabular}{lccc}
\toprule[2pt]
Model & Method & Examples & F1-score \\
\midrule
 &  & 25 & 58.70 \\
RoBERTa$_{\mbox{\scriptsize large}}$(baseline) & Fine-tune & 50 & 66.10 \\
 &  & 100 & 73.78 \\
\midrule
 &  & 25 & 47.33 \\
T5a$_{\mbox{\scriptsize large}}$(baseline) & Fine-tune & 50 & 58.97 \\
 &  & 100 & 70.46 \\
\midrule
 &  & 25 & 57.48 \\
RoBERTa$_{\mbox{\scriptsize large}}$ & MT+MKV & 50 & 71.50 \\
 &  & 100 & 71.85 \\
\midrule
 &  & 25 & \textbf{63.65} \\
T5$_{\mbox{\scriptsize large}}$ & MT+MKV & 50 & \textbf{76.53} \\
 &  & 100 & \textbf{78.01} \\
\bottomrule[2pt]
\end{tabular}
\end{table*}

In our final experiment, we compared the performance of the RoBERTa model, trained using Masked Language Model (MLM) within an Encoder structure, and the T5 model, implemented with an Encoder-Decoder structure, on a resume classification task. Unlike previous model comparison experiments where training was conducted for a fixed four epochs, we adjusted the training duration to the optimal number of epochs based on the test dataset score. This approach better demonstrates the peak performance of both models using the two methods, thereby facilitating a more accurate comparison for determining the more suitable model for the few-shot learning task on the resume dataset.
\par As illustrated in Table \ref{table4}, the T5 model, when trained using the fine-tuning approach, underperforms the RoBERTa model in the 25/50/100-shot outcomes. However, the 25/50/100-shot results outperform the RoBERTa model when the T5 model is trained using the MT+MKV prompt-learning method. Notably, the corresponding improvements are 6.17, 5.03, and 6.16 points respectively.
\par Additionally, the RoBERTa model, when provided with a larger number of training samples, exhibits a performance gain of 1.93 points at the 100-shot level, with the fine-tune method compared to the prompt-learning method. Interestingly, the score difference between the 50-shot and 100-shot instances using the MT+MKV method with the RoBERTa model is minimal, despite doubling the sample size. This suggests limited efficiency in the use of larger samples within this context. This observation aligns with the findings of \cite{lester-etal-2021-power}, which propose that the performance of prompt-tuning improves with an increase in the number of parameters within the model.

\subsection{Analysis of the Confusion Matrix}

\begin{figure}[htbp]
\centering
\subfloat[Knowledgeable Verbalize]{\label{KV}\includegraphics[width=0.45\linewidth]{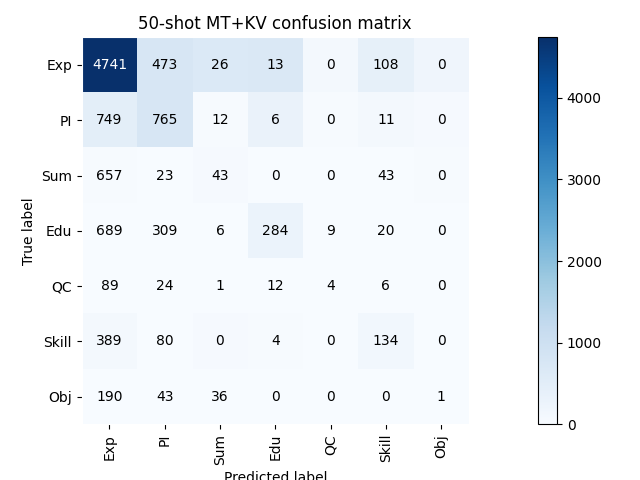}}
\hfill
\subfloat[Manual Knowledgeable Verbalizer]{\label{MKV}\includegraphics[width=0.45\linewidth]{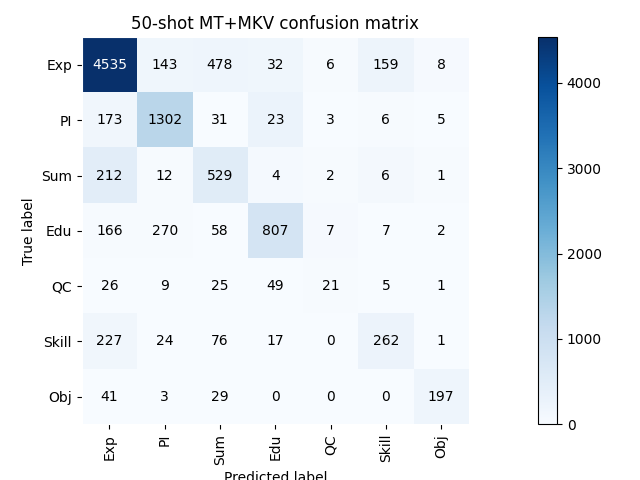}}
\caption{\label{figure5}Confuse Matrix of MT+KV and MT+MKV in 50-shot on T5$_{\mbox{\scriptsize large}}$ model.}

\end{figure}

The “experience” category in a resume dataset features the largest sample size at 41,114, while “qualification” has the smallest at 974, leading to a notable imbalance. This imbalance challenges the use of few-shot learning models. Our proposed MKV method's effectiveness in addressing this imbalance is demonstrated through a confusion matrix comparison in Figure \ref{figure5}.

\par To begin with, an analysis of the confusion matrix for the labels “summary”, “qualification”, “education”, “skill”, and “object” using the original KV method, as depicted in Figure \ref{figure5}(a), reveals a substantial misclassification rate. These labels are associated with a smaller proportion of samples, many of which are erroneously classified into the categories of “experience” and “personal information”, which represent a larger proportion of the sample. Furthermore, the distribution observed in the confusion matrix of the KV method corroborates the trend suggested in Figure \ref{figure5}(a): the fewer the number of samples in a class, the lower the number of correct classifications.
\par Conversely, the application of our proposed MKV method shows a considerable improvement in addressing classification errors associated with sample imbalance, as shown in Figure \ref{figure5}(b). This lends credence to the effectiveness of the MKV approach, crafted according to our rule, in maintaining high performance even in the presence of highly unbalanced samples.

\section{Conclusion}
\label{sec:bibtex}
In this study, we use the prompt technique on the resume dataset for few-shot learning. We created templates informed by resume sentence structures and assessed their utility. Additionally, we refined the construction of a knowledgeable verbalizer, relying on Knowledgeable Prompt Tuning (KPT). For this, we devised construction rules for the MKV, tailored to the textual features of resumes. Experimental evaluations demonstrate our MKV's effectiveness and robustness. While the final outcomes were satisfactory, they also elucidated the constraints inherent in the utilized prompt methodology. It is anticipated that future endeavors will develop a more universal prompt approach, capable of addressing a variety of industries and accommodating diverse resume formats.

\subsubsection{Acknowledgements} This research was  partially supported by JSPS KAKENHI Grant Numbers JP23H00491 and JP22K00502.

\bibliographystyle{splncs04}
\bibliography{References}



%
%
%

%

\end{document}